\documentclass[sigconf]{acmart}

\usepackage{listings}

\AtBeginDocument{%
  \providecommand\BibTeX{{%
    \normalfont B\kern-0.5em{\scshape i\kern-0.25em b}\kern-0.8em\TeX}}}

\setcopyright{acmlicensed}
\copyrightyear{2024}
\acmYear{2024}
\acmDOI{XXXXXXX.XXXXXXX}

\acmConference[Conference acronym 'XX]{Make sure to enter the correct
  conference title from your rights confirmation emai}{April 28--May 2 ,
  2025}{Sydney, Australia}

\acmISBN{978-1-4503-XXXX-X/18/06}

\begin{document}

\title{LLM4GNAS: A Large Language Model Based Toolkit for \\ Graph Neural Architecture Search}


\author{Yang Gao}
\affiliation{%
  \institution{Zhejiang University}
  \city{Hangzhou}
  \country{China}}
\email{gaoyang9507@zju.edu.cn}

\author{Hong Yang*}
\affiliation{%
  \institution{Guangzhou University}
  \city{Guangzhou}
  \country{China}}
\email{hyang@gzhu.edu.cn}

\author{Yizhi Chen}
\affiliation{%
  \institution{Guangzhou University}
  \city{Guangzhou}
  \country{China}}
\email{rbeopad@e.gzhu.edu.cn}

\author{Junxian Wu}
\affiliation{%
  \institution{Guangzhou University}
  \city{Guangzhou}
  \country{China}}
\email{2112333006@e.gzhu.edu.cn}

\author{Peng Zhang}
\affiliation{%
  \institution{Guangzhou University}
  \city{Guangzhou}
  \country{China}}
\email{p.zhang@gzhu.edu.cn}

\author{Haishuai Wang} 
\affiliation{%
  \institution{Zhejiang University}
  \city{Hangzhou}
  \country{China}}
\email{haishuai.wang@zju.edu.cn}

\authornote{Corresponding author: Peng Zhang and Haishuai Wang}

\renewcommand{\shortauthors}{Yang Gao and Peng Zhang, et al.}

\begin{abstract}

Graph Neural Architecture Search (GNAS) facilitates the automatic design of Graph Neural Networks (GNNs) tailored to specific downstream graph learning tasks. However, existing GNAS approaches often require manual adaptation to new graph search spaces, necessitating substantial code optimization and domain-specific knowledge. To address this challenge, we present LLM4GNAS, a toolkit for GNAS that leverages the generative capabilities of Large Language Models (LLMs). LLM4GNAS includes an algorithm library for graph neural architecture search algorithms based on LLMs, enabling the adaptation of GNAS methods to new search spaces through the modification of LLM prompts. This approach reduces the need for manual intervention in algorithm adaptation and code modification. The LLM4GNAS toolkit is extensible and robust, incorporating LLM-enhanced graph feature engineering, LLM-enhanced graph neural architecture search, and LLM-enhanced hyperparameter optimization. Experimental results indicate that LLM4GNAS outperforms existing GNAS methods on tasks involving both homogeneous and heterogeneous graphs.
\end{abstract}



\keywords{Graph Neural Architecture Search, Large Language Models,  Graphs, AutoML, Toolkit. }


\received{20 February 2007}
\received[revised]{12 March 2009}
\received[accepted]{5 June 2009}

\maketitle

\section{Introduction}

Graph Neural Architecture Search (GNAS) \cite{DBLP:conf/ijcai/GaoYZ0H20,gao2022graphnas++} has garnered significant attention as a promising method for automating the design of Graph Neural Networks (GNNs) tailored to downstream graph learning tasks. GNAS aims to alleviate the manual effort required for designing GNN architectures by leveraging reinforcement learning~\cite{gao2022graphnas++}, differential gradient~\cite{DBLP:conf/icde/ZhaoYT21}, and evolutionary algorithm~\cite{Shi2020EvolutionaryAS}. 

However, current GNAS algorithms often face the challenges of designing new search spaces when meeting new downstream graph learning tasks, which calls for heavy manual adjustments to achieve optimal performance. For example, AutoHeG~\cite{zheng2023auto} aims to integrate additional operations of Heterophilic GNNs~\cite{DBLP:journals/corr/abs-2007-02133,bo2021beyond} into traditional search space, which entails not only modifying the search space but also adapting the search algorithm to facilitate the selection of both legacy and novel operators. This limitation hampers the scalability and generalizability of GNAS.

\begin{figure*}
  \includegraphics[width=\textwidth]{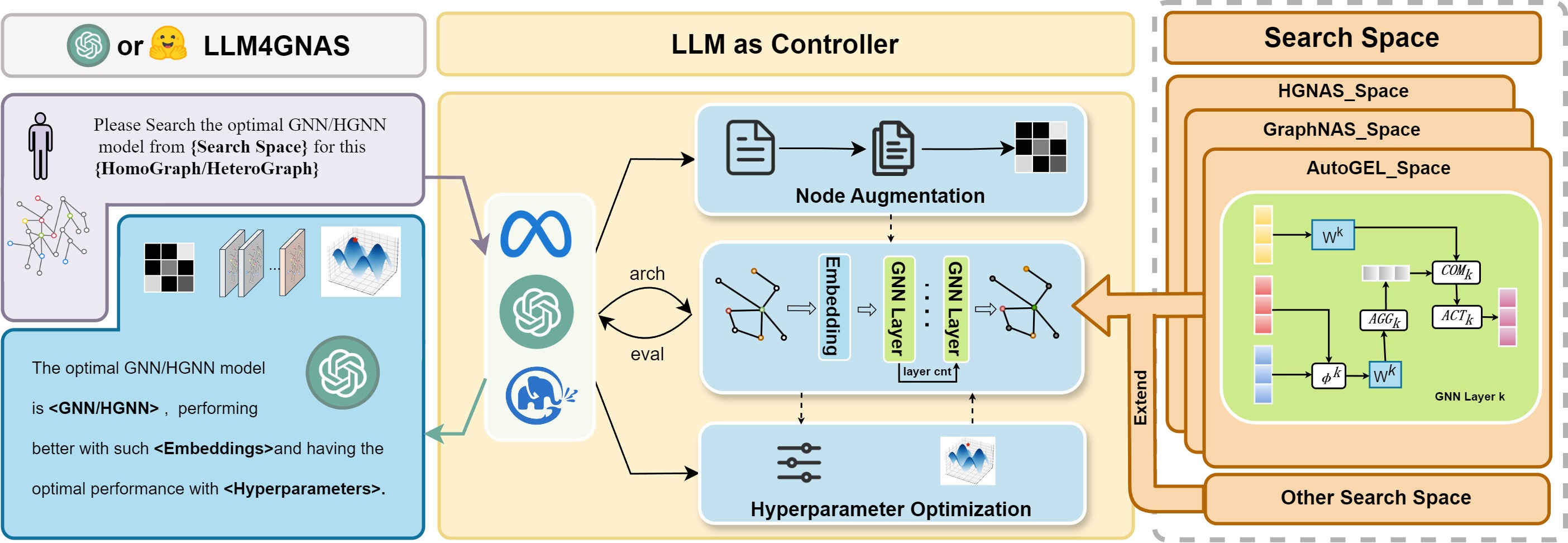}
  \caption{The overall framework of LLM4GNAS.  LLM4GNAS leverages Large Language Models (LLMs) as the controller to automatically design Graph Neural Networks (GNNs) for downstream graph learning tasks. The framework consists of three components, i.e., LLM-based Feature Augmentation, Graph Neural Architecture Search, and Hyperparameter Optimization. Domain knowledge can be injected by  prompts to guide the iterations towards the best architectures.}
  \Description{The overall framework of LLM4GNAS.}
  \label{fig:framework}
\end{figure*}

Recently, Large Language Models (LLMs), such as GPT-4~\cite{openai2023gpt4}, have exhibited remarkable language understanding and generation capabilities. In particular, LLMs have been successfully used in designing novel optimizers~\cite{yang2023large} and CNN architectures~\cite{zheng2023can}, where a new set of prompts are designed  to guide LLMs towards generating better CNN architectures by  learning from prior attempts. Motivated by the success of LLMs in model design and optimization, we aim to integrate LLMs into GNAS by leveraging the generative capabilities of LLMs. Our goal is to develop a new class of GNAS  library based on LLMs that are easily transferable across different  graph tasks and search spaces. 

In this paper, we present LLM4GNAS, a new toolkit for GNAS based on  LLMs.  LLM4GNAS leverages the language understanding and generation capabilities of LLMs to enhance the GNAS process. LLM4GNAS consist of three modules, i.e., LLM-enhanced node augmentation, LLM-enhanced graph neural search algorithms, and LLM-enhanced hyperparameter optimization. First, LLM4GNAS employs LLMs to generate rich features for graph data, which can augment the node features based on LLMs. Second, LLM4GNAS integrates graph neural search algorithms with LLM prompts, which enables diverse exploration of the search space. Third, LLM4GNAS incorporates  hyperparameter optimization driven by LLMs. This approach ensures that the optimal set of hyperparameters can be identified effectively. Our contributions are outlined as follows:
\begin{itemize}
    \item This work introduces a new toolkit, LLM4GNAS, for graph neural architecture search based on LLMs. To our best knowledge, this represents the first work that combines LLMs and graph neural architecture search. 
    \item LLM4GNAS leverages LLMs as the controller and runs  iteratively among three components, i.e., LLM-enhanced Node Augmentation, LLM-enhanced GNAS, and LLM-enhanced hyperparameter Optimization.
    \item Experimental results demonstrate that our method not only achieves competitive results to  existing GNAS tasks, but also enables model  adaptability and scalability in both homogeneous graphs and heterogeneous graphs.
\end{itemize}

We summarize the characteristics of LLM4GNAS as follows. First, \textbf{Easy-to-Use}: LLM4GNAS provides a user-friendly interface. Users can initiate model training on a specific dataset with a single command or a few lines of Python scripts. Second, \textbf{Extensibility}: LLM4GNAS is easily extensible, allowing to add new graph features, search algorithms, and graph encoding functions into the toolkit. Third, \textbf{Rich-Functional}: LLM4GNAS offers a wide range of graph functions, enabling users to perform diverse tasks, such as feature engineering, graph neural architecture search, and hyperparameter optimization based on LLMs.

\section{LLM4GNAS Framework}

In this section, we introduce the framework of LLM4GNAS, as shown in Figure 1. Our framework is built on PyTorch, HuggingFace Transformers, and PYG~\cite{Fey/Lenssen/2019}, which are popular deep learning libraries for graphs and LLMs. 
We first demonstrate how to use LLM as the controller to perform graph neural architecture search, and then introduce three main modules of LLM4GNAS, i.e.,  LLM-enhanced Node Augmentation, LLM-enhanced GNAS, and  LLM-enhanced Hyperparameter Optimization.

\subsection{LLM as Controller}

In LLM4GNAS, LLMs act as controllers of the graph neural architecture search algorithm. The framework begins with the LLM-enhanced Node Augmentation module, which enhances the features of the input graph data by employing LLMs to generate meaningful embeddings for nodes. Following this, LLM4GNAS integrates LLM-based Graph Neural Architecture Search and Hyperparameter Optimization. The Graph Neural Architecture Search utilizes LLMs to explore the search space of possible GNN architectures, aiming to identify architectures and hyperparameters that yield the best performance on the given graph data. The Hyperparameter Optimization module further refines the model's hyperparameters based on performance feedback.

The iterative nature of LLM4GNAS enables continuous refinement of the GNN architecture and its hyperparameters. The performance metrics of the designed model serve as feedback signals, guiding LLM4GNAS in adjusting the architecture and hyperparameters accordingly. This iterative process ensures the creation of highly effective and tailored GNN models capable of addressing diverse graph-related tasks. By incorporating LLMs into the design process, LLM4GNAS enhances the adaptability and performance of GNN models, making it a valuable tool for graph data analysis and processing.

\subsection{ LLM-enhanced Node Augmentation}

In LLM4GNAS, we employ LLM-enhanced node augmentation to generate new node features for downstream tasks. This process incorporates both training-based node augmentation methods, such as TAPE~\cite{he2023harnessing}, and training-free methods that utilize LLMs to generate node embeddings directly.

For labeled graph data, LLM4GNAS adopts explanation-based enhancement methods, such as TAPE~\cite{he2023harnessing} that uses LLMs to generate explanations and pseudo-labels to augment textual attributes. Subsequently, relatively small language models are fine-tuned on both the original text data and the generated explanations to encode text semantic information as initial node embeddings.

For unlabeled graph data, LLM4GNAS employs a feature augmentation approach based on the generation capability of LLMs. By leveraging these generative capabilities, LLM4GNAS can create synthetic nodes that capture the underlying structure of the unlabeled graph data. This feature augmentation method allows the model to extract meaningful representations from the unlabeled data and thus generalizes to unseen graph learning tasks.

\subsection{LLM-enhanced GNAS}

We define the problem of LLM-based graph architecture search as follows. Given an LLM model $\mathbf{\Omega}$, a dataset $\mathcal{G}$, a GNN search space  $\mathcal{M}$, and an evaluation metric $\mathcal{A}$, we aim to find the best architecture $m^{*} \in \mathcal{M}$ on a given graph  $\mathcal{G}$, i.e., 
\begin{equation}
    m^*=\underset{m\in \mathcal{M}(\mathbf{\Omega}(P))}{argmax} \ \ \mathcal{A}(m(\mathcal{G})), 
\end{equation} 
where $\mathcal{M}(\mathbf{\Omega}(P))$ denotes the search space generated by the LLM; $P$ is the GNAS prompt that guides LLM to perform graph neural architecture search; and the metric $\mathcal{A}$, for example, can be either accuracy or AUC for node classification tasks.

The process of LLM-based GNAS involves iteratively designing GNNs and optimizing their architectures based on feedback from their validation performance. This process is analogous to code generation using LLM-based chatbots that adapt according to user feedback. Specifically, we first employ an LLM to design a set of GNN architectures, denoted as $M(\mathbf{\Omega}(P))$. These GNNs are then derived from the search space, constructed, and evaluated on the given graph data. Performance metrics, such as accuracy on the validation set  $A$, serve as feedback signals, guiding the LLM to generate improved GNN architectures. By continuously integrating the generated GNNs $M(\mathbf{\Omega}(P))$ and their evaluation results $A$ in the reward prompt $P_F$, the LLM converges fast. In the last step, GNAS-LLM obtains the best GNN $m^{*}$ generated by the LLM.

We design the GNAS Prompts $P$ to guide LLMs to generate new candidate GNN architectures, i.e., $ \mathcal{M}(\mathbf{\Omega}(P))$. The design of GNAS prompts needs to be aware of the diverse search space and search strategy in GNAS.  Specifically,  GNAS Prompts $P$ contains candidate GNN operations and their connections to describe the search space. Besides, GNAS Prompts $P$ also contains the description of the search strategy to guide the generation of new GNN architectures, such as reinforcement learning-like search strategies used in GNAS-LLM and GHGNAS.

The advantages of LLM4GNAS are summarized as follows.  Firstly, LLM4GNAS enables adaptation to different search spaces by the modification of the GNAS prompts. This capability allows researchers to adjust the architecture search process to suit the characteristics of the target graph data, ensuring that the resulting GNN architecture is well-suited to the specific task. Secondly, LLM4GNAS enables to adjustment of the architecture search process through prompts and enables researchers to iteratively refine the search process, potentially leading to the discovery of more effective architectures.

\subsection{LLM-enhanced Hyperparameter Optimization}

LLM4GNAS also incorporates LLM-enhanced hyperparameter optimization tools. Empirical evaluations~\cite{zhang2023using} indicate that, in settings with constrained search budgets, LLMs can perform comparably to, or even surpass, traditional hyperparameter optimization methods such as random search and Bayesian optimization on standard benchmarks. By leveraging the advanced capabilities of LLMs, LLM4GNAS streamlines the hyperparameter optimization process, thereby reducing both the computational burden and the time required for this essential task.

\section{Interfaces of LLM4GNAS }
In this part, we introduce the interfaces of LLM4GNAS by using different examples. 

\subsection{GNAS Example  with LLM4GNAS}

LLM4GNAS provides a user-friendly interface for designing new GNN models on new graphs. Users can run the program with a single command, as illustrated in the following example:

\noindent
\texttt{python -m llm4gnas.main --search\_space AutoGEL --input /path/data   --task NodeClassification --output /path/result}
\noindent

By specifying the graph data path, search space (e.g., AutoGEL), task type (e.g., Node Classification), and output directory, users can seamlessly configure LLM4GNAS to automatically load the graphs and conduct graph neural architecture search. This streamlined process empowers users to efficiently explore and design GNN models tailored to their specific graph learning tasks.


In addition to supporting command-line training, LLM4GNAS offers Python APIs for user programming, enabling manual datasets loading and automatic GNN design. The following example demonstrates the  usage of LLM4GNAS APIs:

\begin{lstlisting}[language=Python, caption= GNAS Example with LLM4GNAS APIs , label=python_example, frame=single, basicstyle=\ttfamily\small, keywordstyle=\color{blue}, numbers=left, numberstyle=\tiny\color{gray}, numbersep=7pt, breaklines=true]
# load data
from torch_geometric.datasets import Planetoid
dataset = Planetoid("../dataset/", "cora")
# set configuration
config = {
    "dataset_name": "Cora", 
    "taskname": "NodeClassification",
    "in_dim": 2708, "out_dim": 7,
    "llm": "ChatGPT", 
}
# execute GNAS
from llm4gnas.autosolver import AutoSolver
solver = AutoSolver(
    search_space="AutoGEL",
    nas_method="gpt4gnas",
    training_method="global_batch",
    config=config
)
best_gnn = solver.fit(dataset)
labels = solver.predict(dataset, gnn=best_gnn)
\end{lstlisting}

In this example, the dataset is loaded using the Cora dataset from PyTorch Geometric. The GNN architecture search is then performed using the AutoSolver class, with parameters specifying the search space, NAS method, training method, and configuration details. The best GNN model found by the solver is used to predict labels for the dataset.

\subsection{GNAS Example with New Search Space}
LLM4GNAS introduces a new feature that can adapt to new search spaces, making it easy to extend to new graph learning tasks. This feature leverages prompt engineering to adapt, verify, and equip LLM4GNAS for new search spaces, enabling users to integrate their expertise and domain knowledge into the search process. By simplifying the adaptation process through prompts, LLM4GNAS facilitates the exploration of novel search spaces, leading to the development of more effective and specialized GNN architectures for diverse graph tasks. The following example demonstrates how to add a new Search Space to LLM4GNAS:

\begin{lstlisting}[language=Python, caption= GNAS Example with New Search Space , label=python_example_new_search space, frame=single, basicstyle=\ttfamily\small, keywordstyle=\color{blue}, numbers=left, numberstyle=\tiny\color{gray}, numbersep=7pt, breaklines=true]
# Define a new Search Space
from llm4gnas.search_space import SearchSpaceBase
class NewSearchSpace(SearchSpaceBase):
    def __init__(self, config, **kwargs):
        self.config = config
        # Prompts to describe the Search Space
        self.operation_prompt = ""  
        self.connection_prompt = ""  
        self.example_prompt = ""
        ...
    def get_operations(self):
        # Get All candidate operations
        ...
    def to_gnn(self, desc, config):
        # Build GNN with its description and configuration generated by LLM
        ...
# Register with the Model Factory
from llm4gnas.register import model_factory
model_factory["new_search_space"] = NewSearchSpace
# GNAS with new Search Space
from llm4gnas.autosolver import AutoSolver
solver = AutoSolver(
    search_space="new_search_space",
    nas_method="gpt4gnas",
    training_method="global_batch",
    config=config
)
\end{lstlisting}

In this example, we demonstrate how to define and utilize a new search space within the LLM4GNAS framework. This is achieved by creating a new class that inherits from SearchSpaceBase and implementing necessary prompts and methods to describe and construct the GNN architectures based on the new search space. The new search space class needs to be registered with the model factory. After that, LLM4GANS is able to access the new search space via its name in the model factory.

\section{Experiments}

In this section, we present a series of experiments designed to validate the performance of LLM4GNAS. Initially, we evaluate LLM4GNAS on homogeneous graphs. Subsequently, we extend our evaluation to heterogeneous graphs. Finally, we conduct an ablation study to assess the efficiency of LLM4GNAS, complemented by a test case exploring its adaptability to new search spaces.

\subsection{Experiment Setup}

\begin{table}
\centering
\caption{Dataset statistics.}
\label{tab:datasets}
\begin{tabular}{l|rrcc}
\toprule
Dataset & Nodes & Edges & Graphs & Classes \\ \midrule
\multicolumn{5}{c}{Node Classification} \\ \midrule
Cora    & 2,708 & 2,742 & 1 & 7 \\
Pubmed  & 19,717 & 6,258 & 1 & 3 \\
Citeseer & 3,327 & 6,594 & 1 & 6 \\
arXiv   & 169,343 & 1,166,243 & 1 & 40 \\
ogbn products & 54,025 & 74,420 & 1 & 47 \\
tape-arXiv2023 & 46,198 & 78,548 & 1 & 40 \\
ACM & 21,529 & 34,864 & 1 & 3 \\
DBLP & 26,128 & 239,566 & 1 & 4 \\
IMDB & 11,616 & 34,212 & 1 & 3 \\ \midrule
\multicolumn{5}{c}{Graph Classification} \\ \midrule
IMDB-B & 19.8 (Avg.) & 96.53 (Avg.) & 1000 & 2 \\
PROTENS & 39.1 (Avg.) & 72.82 (Avg.) & 1113 & 2 \\
MUTAG & 17.79 (Avg.) & 19.79 (Avg.) & 188 & 2 \\ \midrule
\multicolumn{5}{c}{Link Prediction} \\ \midrule
NS & 1,589 & 2,742 & 1 & 2 \\
Power & 4,941 & 6,594 & 1 & 2 \\
PROTENS & 5,022 & 2,148 & 1 & 2 \\
\bottomrule
\end{tabular}
\end{table}

\begin{figure*}
  \includegraphics[width=\textwidth]{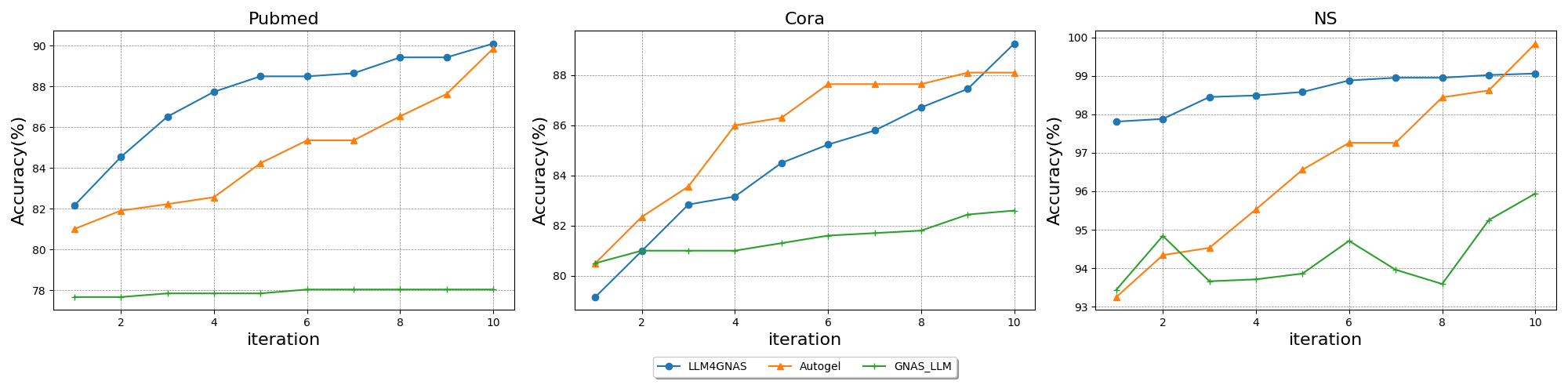}
  \caption{Variation in test accuracy of the best GNNs found at each iteration during the architecture search process for LLM4GNAS, GNAS-LLM, and AutoGEL on homogeneous graphs.}
  \Description{The variation between different method.}
  \label{fig:nc_results}
\end{figure*}

\begin{table*}
\caption{Results on homogeneous graphs.}
\label{tab:gnas}
\begin{tabular}{@{}c|ccc|ccc|ccc@{}}
\toprule
          & \multicolumn{3}{c|}{Node Classification} & \multicolumn{3}{c|}{Link Prediction} & \multicolumn{3}{c}{Graph Classification} \\ 
Method    & Cora       & Citeseer      & Pubmed     & NS       & Power      & Router      & IMDB-B      & MUTAG      & PROTEINS      \\ \midrule
GCN       & 87.63±0.20      & 76.66±0.60         & 88.58±0.45      & 92.70±1.63         & 70.10±0.12           & 70.30±0.10            & 74.00±1.60            & 85.60±1.56           & 76.00±0.26              \\
GAT       & 86.09±0.33      & 75.66±0.57         & 88.34±2.00      & 91.05±0.33         & 66.08±1.56           & 64.02±0.56            & \textbf{88.23±0.04}            & 72.40±0.56           & 75.71±0.01              \\
GraphSAGE & 87.41±0.58           & 75.99±1.30              & 88.34±0.49           & 94.89±0.57         & 67.63±1.30           & 75.45±0.66            & 72.30±0.69            & 85.10±1.23           & 75.90±0.60              \\
GCNII     & 85.50±0.50           & 73.40±0.60              & 80.30±0.40           & 92.33±0.26         & 82.77±0.45           & 77.85±0.60            & 71.62±0.33            & 71.57±0.66           & 70.13±0.37               \\
GATv2     & 79.30±1.50           & 70.65±0.80              & 79.80±1.86           & 90.87±1.33         & 70.55±0.76           & 63.22±0.40            & \underline{81.57±0.60}            & 56.57±0.31           & 78.02±0.28              \\ \midrule
Random    & 86.98±0.11           & 72.98±0.78              & 86.62±0.30           & 90.86±1.55         & 69.33±1.55           & 74.96±0.73            & 72.00±0.20            & 83.40±0.50           & 74.11±1.30              \\
GraphNAS  & 88.90±0.56           & 77.60±2.30              & \textbf{91.2±1.25}           & -         & -           & -            & -            & -           & -              \\
GNAS-LLM  & 91.51±0.66      & \textbf{78.08±0.50}         & 89.62±1.45      & 91.24±1.20         & 72.24±1.55           & 85.66±1.33            & 72.50±0.10            & 81.57±0.31           & 72.19±0.65              \\
AutoGEL   & \textbf{89.89±0.62}      &  \underline{77.66±0.33}         & 89.68±1.25      &  \underline{99.86±0.06}         &  \textbf{98.00±0.21}           & \underline{99.08±0.28}            & 81.20±0.20            & \textbf{94.74±1.84}           & \textbf{82.68±0.64}              \\
\textbf{LLM4GNAS}  & \underline{89.2±0.63}       & 76.2±0.66          & \underline{90.3±1.30}       & \textbf{99.88±1.20}         & \underline{91.88±1.85}           & \textbf{99.33±0.10}            & 76.00±0.30            & \underline{94.73±1.20}           & \underline{76.23±0.52}              \\ \bottomrule
\end{tabular}
\end{table*}

\subsubsection{Baselines}
We compare GNNs designed by LLM4GNAS with the most popular GNNs, including GCN~\cite{kipf2016semi}, GAT~\cite{velickovic2017graph}, GCNII, and GATv2~\cite{brody2021attentive}. For experiments on heterogeneous graphs, we  take RGCN~\cite{schlichtkrull2018modeling}, HAN~\cite{wang2019heterogeneous}, HGT~\cite{hu2020heterogeneous}, MAGNN~\cite{fu2020magnn} as additional baselines.  We also compare the proposed model with other GNAS methods, including  Random Search, GraphNAS~\cite{DBLP:conf/ijcai/GaoYZ0H20}, HGNAS~\cite{gao2021heterogeneous},  and AutoGEL~\cite{wang2021autogel}. We use the same experimental setting used in AutoGEL for homogeneous graphs and HGNAS++~\cite{gao2023hgnas++} for heterogeneous graphs.
    
\subsubsection{Hyperparameters}
LLM4GNAS runs 15 search iterations. The number of iterations is constrained by the length of prompts  that the LLMs allow. During each iteration, LLMs generate 10 new architectures. For each architecture search, we only use one candidate operation connections and nine candidate operations.
In the following experiments, we repeat our method three times and show the best results \textit{w.r.t.} accuracy on a validation dataset.

If not otherwise specified, we use GPT-4 with version V20230314 as the default LLM in the experiments. For all models, we set temperature $\tau = 0$ for reproducibility. We adopt accuracy as the metric for all tasks. Additionally, we use all the GNAS baselines to generate $N=10$  architectures at each iteration. Specifically, dor GraphNAS, the ADAM optimizer is used, with a learning rate of 0.00035.
In the experiments involving AutoGEL, we set the layer number to 2, the ADAM optimizer with a learning rate of 5e-4, a minibatch size of 128, and train each generated architecture for 200 epochs with a dropout value of 0.5.

\subsubsection{Experimental setting}
The dataset statistics are presented in Table \ref{tab:datasets}. For experiments conducted on homogeneous graphs, we adopt the identical experimental settings used in AutoGEL to ensure a fair comparison. Similarly, for experiments involving heterogeneous graphs, we utilize the same experimental settings as those reported in HGNAS++ \cite{gao2023hgnas++} to maintain comparability.

\subsubsection{Experimental environment}  We run the experiments on Ubuntu 18.04 with Python 3.8, PyTorch 1.12.1+cu113, PyTorch Geometric 2.3.1, scikit-learn 1.2.2, Numpy 1.24.3,  and NAS-Bench-Graph 1.4.0.

\subsection{Results on Homogeneous Graphs}

We evaluated the performance of LLM4GNAS on homogeneous graphs across node classification, link prediction, and graph classification tasks, adhering to the same experimental settings as used in AutoGEL.

\begin{table*}
\caption{Results on heterogeneous Graphs.}
\label{tab:hgnas}
\begin{tabular}{@{}c|ccc|ccc@{}}
\toprule
         & \multicolumn{3}{c}{Node Classification|} & \multicolumn{3}{c}{Link Prediction} \\
Method   & ACM         & DBLP        & IMDB        & Amazon     & Yelp    & MovieLens    \\ \midrule
GCN      &90.33±0.59   &83.03±0.93  &55.19±0.99   &78.12±0.06 &91.62±0.11 &90.21±0.32              \\
GAT      &91.25±0.27   &84.34±0.94  &53.37±1.27   &75.43±0.07 &85.51±0.20 &87.38±0.46                      \\
GCNII    &88.07±0.12   &80.18±0.92  &54.59±0.08   &75.05±0.53  &79.93±0.03 &87.72±0.10              \\
GATv2    &90.74±0.18   &82.76±0.11  &56.84±0.05   &77.65±0.37  &89.73±0.51 &88.20±0.17              \\
RGCN     &91.25±0.50   &93.66±0.76  &55.05±0.12   &73.85±0.12  &88.63±0.17 &85.25±0.30      \\
HAN      &91.16±0.27   &92.12±0.20  &55.59±0.67   &75.96±0.23  &89.64±0.21 &86.46±0.13              \\
HGT      &88.93±0.38   &92.97±0.20  &59.35±0.79   &74.30±0.02  &91.73±0.20 &87.93±0.12              \\
MAGNN    &91.30±0.38   &93.57±0.20  &56.44±0.63   &75.90±0.33  &90.50±0.30 &86.03±0.35             \\ \midrule
Random   &90.89±0.74   &93.24±2.13  &57.12±0.42   &77.52±1.37  &90.89±0.74 &89.71±0.60              \\
HGNAS    &91.21±0.63   &91.35±1.36  &57.32±0.54   &\underline{78.50±0.39}  &\underline{91.96±0.21} &\underline{90.63±0.33}              \\
DiffMG   &\underline{92.65±0.15}   &\textbf{94.45±0.15}  &\underline{61.04±0.56}   & -  & -   & -            \\

\textbf{LLM4GNAS} &\textbf{92.97±0.61}  &\underline{94.41±0.35}  &\textbf{62.66±1.72}    &\textbf{79.04±0.16}  &\textbf{92.92±0.14}  &\textbf{92.76±0.12}             \\ \bottomrule
\end{tabular}
\end{table*}

\begin{figure*}
  \includegraphics[width=\textwidth]{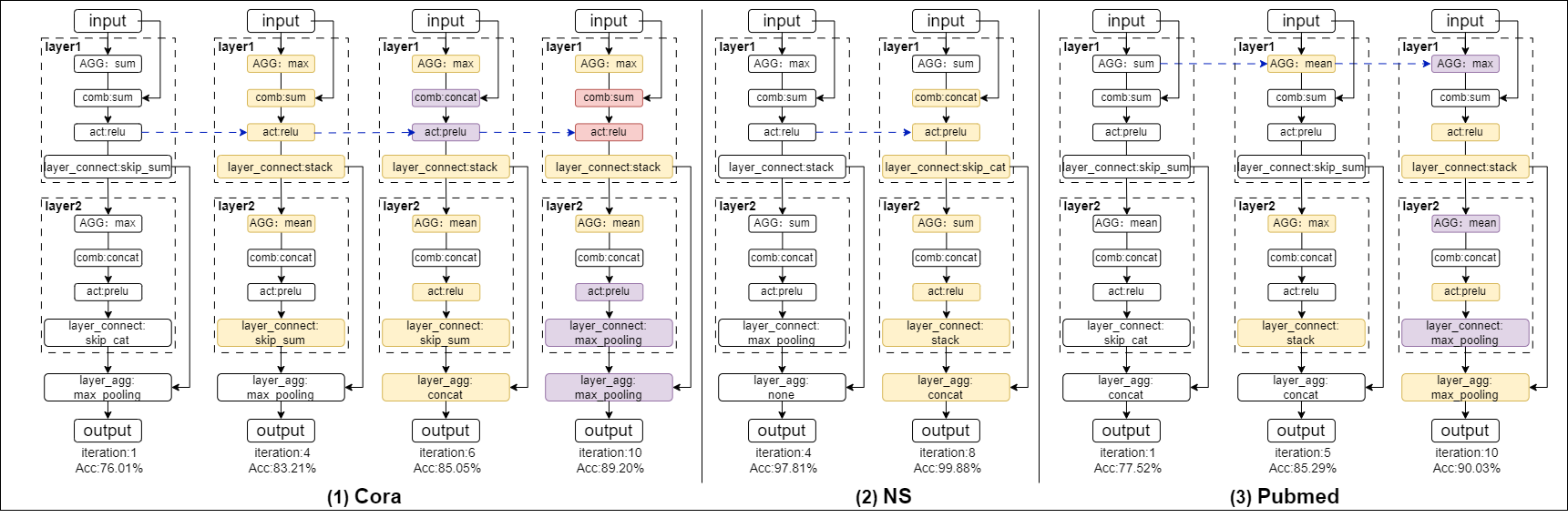}
  \caption{Iterations of the best GNN architectures generated by LLM4GNAS on homogeneous graphs. Colored blocks indicate modifications from the previous iteration.}
  \Description{The iteration of LLM4GNAS in different dataset.}
  \label{fig:architects}
\end{figure*}

Table~\ref{tab:gnas} presents a comparison of LLM4GNAS with several baseline models across various tasks on homogeneous graphs. Specifically, in link prediction tasks, LLM4GNAS attains accuracies of 99.88\% on the NS dataset and 99.33\% on the Router dataset, outperforming the other methods. Our method performs worse than AutoGEL on the Power dataset but higher than the other baseline models. For node classification tasks, LLM4GNAS yields accuracies of 89.20\% on Cora and 90.30\% on PubMed, ranking second on both datasets; AutoGEL and GraphNAS achieve the highest accuracies on Cora (89.89\%) and PubMed (91.20\%), respectively. On Citeseer, LLM4GNAS obtains an accuracy of 76.20\%, which is comparable to other methods but slightly lower than the best-performing models. In graph classification tasks, LLM4GNAS achieves an accuracy of 94.73\% on MUTAG, closely matching AutoGEL's result of 94.74\%. On the PROTEINS dataset, it attains 76.23\%, representing the second-highest performance among the compared methods. Although LLM4GNAS does not achieve the highest accuracy on IMDB-B, it maintains competitive performance.

To further illustrate the effectiveness of LLM4GNAS during the architecture search process, we present Figure~\ref{fig:nc_results} and Figure~\ref{fig:architects}. Figure~\ref{fig:nc_results} depicts the variation in test accuracy of the best GNN architectures identified at each iteration by GNAS methods, where LLM4GNAS demonstrates a rapid improvement over the initial iterations, consistently surpassing the performance of the other methods. Figure~\ref{fig:architects} shows the iterations of the best GNN architectures generated by LLM4GNAS on homogeneous graphs, highlighting the modifications made at each step. These figures indicate that the integration of LLMs enables LLM4GNAS to more effectively explore the search space, leading to the identification of superior architectures in fewer iterations. Consequently, LLM4GNAS not only enhances the performance of the resulting GNN models but also improves the efficiency and stability of the architecture search process.

\subsection{Results on Heterogeneous Graphs}

We evaluated LLM4GNAS on heterogeneous graphs for node classification and link prediction tasks, following the experimental setup used in HGNAS++. The node classification experiments were conducted on the ACM, DBLP, and IMDB datasets, while the link prediction tasks were performed on the Amazon, Yelp, and MovieLens datasets. The results are summarized in Table~\ref{tab:hgnas}.

In both node classification and link prediction tasks, LLM4GNAS demonstrated strong performance. For node classification, it achieved the highest accuracies on the ACM dataset (92.97\%) and the IMDB dataset (62.66\%). On the DBLP dataset, LLM4GNAS attained an accuracy of 94.41\%, which is slightly below the highest score obtained by DiffMG (94.45\%) but higher than all other compared methods. Compared to baseline models such as GCN, GAT, GCNII, and others, LLM4GNAS consistently showed improved results. In the link prediction tasks, LLM4GNAS outperformed all other methods on the Amazon, Yelp, and MovieLens datasets, achieving accuracies of 79.04\%, 92.92\%, and 92.76\%, respectively. These results surpass those of HGNAS, which previously held the highest accuracies among the compared models for these datasets.

The comparative analysis indicates that LLM4GNAS effectively leverages LLMs to explore the architecture search space in heterogeneous graphs, resulting in models that perform better on both node classification and link prediction tasks. The consistent performance gains across multiple datasets suggest that LLM4GNAS is robust and generalizes well to various heterogeneous graph settings.

\begin{table}[]
\caption{Result of Ablation study on Node Augmentation.}
\label{tab:aug}
\begin{tabular}{@{}lllll@{}}
\toprule
Model        & Cora                                                                 & Citeseer                                                           & Pubmed                                                            & Arxiv                                                             \\ \midrule
GCN          & \begin{tabular}[c]{@{}l@{}}\underline{88.01±0.21}\end{tabular}    & \begin{tabular}[c]{@{}l@{}} 68.90±0.01\end{tabular}      & \begin{tabular}[c]{@{}l@{}}83.62±0.03\end{tabular} & \begin{tabular}[c]{@{}l@{}}68.43±0.26\end{tabular} \\
GCN+NA    & \begin{tabular}[c]{@{}l@{}}87.45±0.01\end{tabular} & \begin{tabular}[c]{@{}l@{}}73.80±0.01\end{tabular}  & \begin{tabular}[c]{@{}l@{}}\underline{93.53±0.15}\end{tabular} & \begin{tabular}[c]{@{}l@{}}74.19±0.04\end{tabular} \\ \midrule
GAT          & \begin{tabular}[c]{@{}l@{}}83.72±0.31\end{tabular}    & \begin{tabular}[c]{@{}l@{}}65.9±0.01\end{tabular}     & \begin{tabular}[c]{@{}l@{}}83.29±0.03\end{tabular} & \begin{tabular}[c]{@{}l@{}}76.50±0.02\end{tabular} \\ 
GAT+NA   & \begin{tabular}[c]{@{}l@{}}87.82±0.01\end{tabular}    & \begin{tabular}[c]{@{}l@{}}\underline{75.00±0.12}\end{tabular} & \begin{tabular}[c]{@{}l@{}}89.10±0.01\end{tabular} & \begin{tabular}[c]{@{}l@{}}\underline{79.19±0.26}\end{tabular} \\ \midrule
GCNII        & \begin{tabular}[c]{@{}l@{}}\textbf{89.11±0.25}\end{tabular}    & \begin{tabular}[c]{@{}l@{}}66.20±0.01\end{tabular}     & \begin{tabular}[c]{@{}l@{}}87.37±0.15\end{tabular} & \begin{tabular}[c]{@{}l@{}}69.85±0.33\end{tabular} \\
GCNII+NA & \begin{tabular}[c]{@{}l@{}}87.82±0.22\end{tabular}    & \begin{tabular}[c]{@{}l@{}}\textbf{75.80±0.41}\end{tabular} & \begin{tabular}[c]{@{}l@{}}\textbf{94.37±0.43}\end{tabular} & \begin{tabular}[c]{@{}l@{}}\textbf{81.57±0.23}\end{tabular} \\ \bottomrule
\end{tabular}
\end{table}

\subsection{Ablation Study}

We design ablation experiments to verify that the design of LLM4GNAS are effective. Firstly, we design ablation experiments on modules of LLM4GNAS to verify that LLM-based Node Augmentation and HPO are effective. Furthermore, we design ablation experiments on LLMs to show our toolkit can adapt to different LLMs. Lastly, we design a case study on a new search space to show  our toolkit can adapt to a new search space.

\subsubsection{On Modules}

We investigated the impact of incorporating LLM-enhanced node augmentation into various graph neural networks (GNNs) to generate more informative node embeddings for downstream tasks. The node augmentation process enriches the feature space of graph nodes by integrating external data, specifically utilizing large language models (LLMs) to supplement the textual information associated with nodes. By introducing additional contextual features, this approach allows models to capture more nuanced node representations, potentially leading to improved performance in tasks such as node classification. We integrated this augmentation technique with several GNN architectures, including GCN, GAT, and GCNII, which traditionally rely solely on the structural and inherent feature information present in the dataset.

As shown in Table \ref{tab:aug}, incorporating node augmentation resulted in performance improvements across multiple datasets, particularly those rich in features. For example, on the PubMed dataset, GCN with node augmentation (GCN+NA) improved accuracy from 83.62\% to 93.53\%. Similarly, on the Arxiv dataset, GCNII+NA achieved an accuracy of 81.57\%, compared to 69.85\% with GCNII alone, demonstrating the benefit of the added node augmentation. However, the performance gains were not uniform across all datasets and models. In the Cora dataset, the addition of node augmentation to the GCN model did not significantly enhance results. Conversely, in more challenging datasets like Citeseer, GAT+NA outperformed the baseline GAT model, increasing accuracy from 65.9\% to 75.0\%. Additionally, GCNII+NA showed consistent improvements across all datasets, achieving the highest performance of 75.80\% in Citeseer, surpassing both GCN and GAT variants with node augmentation. These findings suggest that the effectiveness of node augmentation depends on the dataset characteristics and that combining a robust model architecture with enriched node features can lead to substantial performance enhancements, especially in tasks where the structural complexity of the graph requires advanced feature extraction.

\subsubsection{On Different LLMs}

We evaluated the impact of using different large language models (LLMs) within our LLM4GNAS framework by employing GPT-4, GPT-3, LLAMA-2, and GLM-3. The results of this ablation study are presented in Table \ref{tab:LLMs}. LLM4GNAS utilizing GPT-4 achieved the highest accuracy across all datasets. On the Cora dataset, for instance, LLM4GNAS with GPT-4 attained an accuracy of 89.22\%, compared to 88.90\% with GPT-3, 87.88\% with GLM-3, and 85.45\% with LLAMA-2. Similar patterns were observed on the Citeseer and Pubmed datasets. These results indicate that the choice of LLM has a significant effect on the performance of our method, with GPT-4 providing the most effective enhancements.

\begin{table}[tb]
\caption{Result of Ablation study on different LLMs.}
\label{tab:LLMs}
\begin{tabular}{@{}llll@{}}
\toprule
Method  & Cora                                                                       & Citeseer                                                                  & Pubmed                                                                   \\ \midrule
GPT-4   & \begin{tabular}[c]{@{}l@{}}\textbf{89.22}\end{tabular} & \begin{tabular}[c]{@{}l@{}}\textbf{76.25}\end{tabular} & \begin{tabular}[c]{@{}l@{}}\textbf{90.30}\end{tabular} \\            
GPT-3   & \begin{tabular}[c]{@{}l@{}}\underline{88.90}\end{tabular} & \begin{tabular}[c]{@{}l@{}}75.90\end{tabular} & \begin{tabular}[c]{@{}l@{}}88.40\end{tabular} \\

LLAMA-2 & \begin{tabular}[c]{@{}l@{}}85.45\end{tabular} & \begin{tabular}[c]{@{}l@{}}\underline{75.39}\end{tabular} & \begin{tabular}[c]{@{}l@{}}\underline{88.73}\end{tabular} \\

GLM-3  & \begin{tabular}[c]{@{}l@{}}87.88\end{tabular} & \begin{tabular}[c]{@{}l@{}}73.63\end{tabular} & \begin{tabular}[c]{@{}l@{}}86.43\end{tabular} \\ \bottomrule
\end{tabular}
\end{table}

\subsubsection{On New Search Space}

We evaluated LLM4GNAS within the search space defined by NAS-Bench-Graph, using the same experimental settings outlined in that benchmark. This approach ensures fair comparisons and consistency across all evaluated models. The primary objective was to assess the architecture search capabilities of LLM4GNAS relative to other established methods such as Random search, GraphNAS, and Genetic-GNN, all of which have been previously tested within the same search space.

As shown in Table 6, LLM4GNAS outperformed its counterparts across various datasets, achieving the highest accuracy in all tasks. For instance, on the Cora dataset, LLM4GNAS attained an accuracy of 80.93\%, surpassing the results of Genetic-GNN and GraphNAS. Similarly, on the Citeseer dataset, it achieved an accuracy of 70.22\%, slightly exceeding other models. In the Pubmed dataset, LLM4GNAS achieved 79.87\% accuracy, outperforming both GraphNAS and Random methods. Additionally, on the Arxiv dataset, it matched the top performance of Genetic-GNN with an accuracy of 73.41\%. These results indicate that LLM4GNAS is effective in navigating the architecture search space, consistently achieving high accuracy across diverse graph datasets. This suggests its potential utility as a tool for neural architecture search in graph-based learning tasks.

\begin{small}
\begin{table}[htb]
  \caption{Performance of three-time architecture search \textit{w.r.t.} Accuracy(\%) on NASBench-Graph.}
  \label{tab:average3}
  \centering
  \begin{tabular}{c|c|c|c|c}
    \toprule
    Methods & Cora & Citeseer & Pubmed & Arxiv \\
    \midrule
    Random & 79.69$\pm$0.31 & 69.96$\pm$0.37 & \underline{79.49$\pm$0.91} & 73.31$\pm$0.04 \\
    \midrule
    GraphNAS & 79.69$\pm$0.20 & \underline{70.11$\pm$0.21} & 79.36$\pm$0.84 & 73.33$\pm$0.17 \\
    \midrule
    Genetic-GNN & \underline{80.27$\pm$0.68} & 70.02$\pm$1.00 & 78.89$\pm$1.00 & \underline{73.41$\pm$0.13} \\
    \midrule
    \textbf{LLM4GNAS} & \textbf{80.93$\pm$0.00} & \textbf{70.22$\pm$0.27} & \textbf{79.87$\pm$0.46} & \textbf{73.41$\pm$0.07} \\
    \bottomrule
  \end{tabular}
\end{table}
\end{small}

\subsubsection{On New Graphs}

We conduct experiment on a large graph dataset, ogbn-products~\cite{hu2020open}, which contains 2,449,029 nodes and 61,859,140 edges. Each node has 100 features, and a total of 47 different node classes.
 We use the functions provided by the SGL library~\cite{zhang2022pasca}. For the search process, we use GNAS-LLM to search 10 architectures in each of the 15 rounds. We also use random search and Pasca~\cite{zhang2022pasca} as baselines, exploring a total of 150 architectures. Ultimately, our method achieves an accuracy of 70.86\% on the test set, which is 0.31\% higher than 70.55\% achieved by the random search and 1.36\% higher than 69.50\% achieved by the Pasca.
 
 The case study, we compared our approach with the baseline in the search space specifically designed for large graphs, and our method achieved leading results. This demonstrates that our approach can also handle the architecture search of large-scale graphs.

\section{Related Work}

\subsection{Graph Neural Architecture Search (GNAS)} 

GraphNAS~\cite{DBLP:conf/ijcai/GaoYZ0H20} is among the earliest methods employing reinforcement learning to design GNN architectures. Building upon GraphNAS, AutoGNN~\cite{AutoGNN} introduces an entropy-driven candidate model sampling method and a weight-sharing strategy to select GNN components more efficiently. GraphNAS++~\cite{gao2022graphnas++} accelerates GraphNAS by utilizing distributed architecture evaluation. GM2NAS~\cite{gao2023gm2nas} also employs reinforcement learning to design GNNs for multitask multiview graph learning, while MVGNAS~\cite{al2022multi} is tailored for biomedical entity and relation extraction. Additionally, HGNAS~\cite{DBLP:conf/icdm/GaoZLZLH21} and HGNAS++~\cite{gao2023hgnas++} utilize reinforcement learning to discover heterogeneous graph neural networks.

In contrast to reinforcement learning-based GNAS methods that explore a discrete GNN search space, differentiable gradient-based GNAS methods have emerged, exploring a relaxed, continuous GNN search space. These methods include DSS~\cite{li2021one}, SANE~\cite{DBLP:conf/icde/ZhaoYT21}, GAUSS~\cite{guan2022large}, GRACES~\cite{qin2022graph}, AutoGT~\cite{zhang2022autogt}, and Auto-HEG~\cite{zheng2023auto}. SANE focuses on discovering data-specific neighborhood aggregation architectures, while DSS is designed for GNN architectures with a dynamic search space. GAUSS addresses large-scale graphs by devising a lightweight supernet and employing joint architecture-graph sampling for efficient handling. GRACES aims to generalize under distribution shifts by tailoring GNN architectures to each graph instance with an unknown distribution. AutoGT extends GNAS to Graph Transformers, and Auto-HEG enables automated graph neural architecture search for heterophilic graphs. Furthermore, methods such as AutoGEL~\cite{wang2021autogel}, DiffMG~\cite{diffmg}, MR-GNAS~\cite{Zheng2022MultiRelationalGN}, and DHGAS~\cite{Zhang_Zhang_Wang_Qin_Qin_Zhu_2023} focus on heterogeneous graphs.

Evolutionary algorithms have also been employed in GNAS by methods such as AutoGraph~\cite{li2020autograph} and Genetic-GNN~\cite{Shi2020EvolutionaryAS}, which aim to identify optimal GNN architectures. G-RNA~\cite{DBLP:conf/cvpr/XieCZWWZY023} introduces a unique search space and defines a robustness metric to guide the search process for defensive GNNs. Surveys by Zhang et al.~\cite{DBLP:conf/ijcai/ZhangW021} and Oloulade et al.~\cite{oloulade2021graph} provide comprehensive overviews of automated machine learning methods on graphs.

GNAS-LLM~\cite{wang2023graph} and GHGNAS~\cite{dong2023heterogeneous} explore the application of LLMs to enhance the GNAS search process. Building upon these works, we introduce an LLM-based GNAS toolkit that integrates feature augmentation, graph neural architecture search, and hyperparameter optimization driven by LLMs. This approach allows our toolkit to adapt to complex search spaces and design GNN architectures for new tasks, thereby advancing the flexibility of GNAS methodologies.

\subsection{Large Language Models}

GPT-4 has emerged as an AI model capable of providing responses to inquiries involving multi-modal data~\cite{openai2023gpt4}. Studies indicate that GPT-4 exhibits proficiency in comprehending graph data~\cite{guo2023gpt4graph}, demonstrating strong performance across various graph learning tasks. Additionally, research has integrated large language models with graph learning models, utilizing GPT-4 for reasoning over graph data~\cite{zhanggraph}. In contrast, BERT adopts a pre-training approach on extensive unlabeled data, followed by fine-tuning on specific downstream tasks~\cite{DBLP:conf/naacl/DevlinCLT19}. Building upon BERT, several language models have been proposed~\cite{DBLP:journals/corr/abs-1907-11692,DBLP:conf/iclr/LanCGGSS20,DBLP:conf/iclr/HeLGC21}. For instance, PaLM is constructed based on the decoder of Transformers~\cite{Vaswani2017AttentionIA}, with PaLM 2 achieving improved results through a larger dataset and a more intricate architecture~\cite{chowdhery2022palm,anil2023palm}.

Recent developments have seen the use of large language models for neural architecture search. GENIUS, for example, leverages GPT-4 to design neural architectures for CNNs, exploring the generative capacity of LLMs~\cite{zheng2023can}. The fundamental concept involves enabling GPT-4 to learn from feedback on generated neural architectures, thereby iteratively improving the design. Experimental results on various benchmarks demonstrate GPT-4's ability to discover top-ranked architectures after several prompt iterations. AutoML-GPT introduces a series of prompts for LLMs to autonomously undertake tasks such as data processing, model architecture design, and hyperparameter tuning~\cite{zhang2023automl}. Notably, GPT-4 has been introduced into graph neural architecture search by GNAS-LLM~\cite{wang2023graph} and GHGNAS~\cite{dong2023heterogeneous}, aiming to search for the optimal GNN or heterogeneous GNN within a designated search space.

In this paper, we present a toolkit that extends the application of LLMs to the generation of new GNN architectures. The toolkit is designed to be easily extendable to new search spaces and graph tasks, offering a user-friendly interface for practical applications and academic research. Our aim is to enhance the performance of existing GNAS methods by providing a versatile and efficient solution for GNN design.

\section{Conclusions}

In this work, we introduced LLM4GNAS, a toolkit designed to automatically generate GNN models tailored to a variety of graph-related tasks. The toolkit integrates LLM-based node augmentation methods, GNAS strategies, and hyperparameter optimization techniques. By leveraging the capabilities of LLMs, LLM4GNAS effectively extends the search space and accommodates different types of graphs, demonstrating adaptability and efficacy in enhancing GNN performance.
LLM4GNAS contributes to the field of GNN architecture search by providing a versatile solution capable of addressing a range of graph-related challenges. The integration of LLMs within the toolkit automates and optimizes the design process, aiming to achieve high performance and generalization across multiple tasks.

Future work will focus on enhancing the toolkit by automating the design of search strategy prompts and developing transferable LLM-based GNAS methods specifically for graph foundation models. By concentrating on the architecture design and fine-tuning of these graph foundation models, we aim to streamline the neural architecture search process and improve the adaptability of LLM4GNAS to a broader spectrum of graph-related problems. These efforts are intended to further optimize the toolkit's effectiveness and extend its applicability across diverse graph learning tasks.

\newpage
\balance
\bibliographystyle{plain}
\bibliography{sample-base}

\end{document}